\documentclass[runningheads]{llncs}
\usepackage[T1]{fontenc}
\usepackage[utf8]{inputenc}
\usepackage{bm}
\usepackage{esvect}

\usepackage{graphicx,verbatim}
\usepackage{nicematrix}  
\usepackage[table]{xcolor}  
\usepackage{booktabs}
\usepackage{lipsum}
\usepackage{duckuments}
\usepackage{graphicx}
\usepackage{subcaption}
\usepackage{booktabs}
\usepackage[table]{xcolor}
\usepackage{amssymb}
\usepackage{multirow}
\usepackage{cite}
\usepackage{hyperref}

\definecolor{MICCAIBlue}{HTML}{0498BB}
\definecolor{DarkBlue}{HTML}{002453}
\definecolor{LightBlue}{HTML}{8CCFD9}
\definecolor{PaleBlue}{HTML}{DDF2F5}

\begin{document}
\title{Evaluating the Generalization of Neuroimaging Foundation Models on African Brain MRI}

\titlerunning{Evaluating the Generalization of Neuroimaging Foundation Models}

\author{
Oluwatobi Iyanuoluwa Akinmuleya\thanks{Equal contribution}\inst{1} \and
Olatokun Shamsudeen Akano\protect\footnotemark[1]\inst{2} \and
Samuel Danquah Ankapong\inst{3} \and
Olamide Lawal\inst{4} \and
Toufiq Musah\inst{5}
}

\authorrunning{Akinmuleya et al.}

\authorrunning{Akinmuleya et al.}

\institute{
Department of Medicine, Shenyang Medical College, Shenyang, China
\and
College of Medicine, University of Ibadan, Ibadan, Nigeria
\and
University of Ghana, Accra, Ghana
\and
Obafemi Awolowo University, Ile-Ife, Nigeria
\and
Kwame Nkrumah University of Science and Technology, Kumasi, Ghana
}

\maketitle              % typeset the header of the contribution

\begin{abstract}
Neuroimaging foundation models pretrained on large, predominantly western cohorts are increasingly proposed as general-purpose backbones for brain MRI analysis. Yet, their ability to generalize to underrepresented clinical populations remains largely untested. We evaluate four recent foundation models (BrainIAC, Neuro-JEPA, NeuroVFM, and Primus) on a three-way diagnostic classification task (Control, Dementia, Parkinson's disease) using a cohort of 88 subjects from a Nigerian clinical brain MRI dataset, across four modality configurations (T1w, T2w, T1w+T2w, FLAIR), and compare against an end-to-end trained ViT3D baseline. The frozen backbones collapse to majority-class predictions, while Neuro-JEPA on FLAIR shows modest but still limited discrimination. In contrast, the end-to-end trained ViT3D achieves higher accuracy and MCC on every task (up to 53.4\% accuracy, MCC=0.27) and is the only model with non-trivial recall. Our findings suggest that these frozen neuroimaging foundation models are insufficient for fine-grained diagnostic classification in small, non-western clinical cohorts, motivating parameter-efficient adaptation and broader multi-site external validation for equitable deployment in global health settings.

% Authors must provide keywords and are not allowed to remove this Keyword section.

\end{abstract}

\section{Introduction}
 Diagnostic infrastructure for brain imaging remains critically underdeveloped across the continent \cite{wogu2025fair, ogbole2018survey}. Fewer than 1\% of global dementia research data originate from Africa \cite{oyeniran2025imaging, akinyemi2025epidemiology}. At the same time, brain MRI foundation models
are increasingly pretrained on large datasets drawn predominantly from well-resourced, non-African healthcare systems, and positioned as general-purpose representations for downstream clinical tasks. Whether these representations generalize to the heterogeneous acquisition conditions and clinical populations encountered in African settings therefore remains unclear.

Automated detection of neurodegenerative disease has become an active application area for AI in neuroimaging \cite{chandru2025artificial}. Alzheimer's disease, Parkinson's disease, and other dementia subtypes share overlapping structural signatures, including cortical and subcortical atrophy patterns that vary gradually and unevenly across regions \cite{caligiore2022neurodegenerative}. These overlapping patterns can make early or borderline cases difficult to distinguish from routine clinical
imaging alone. Machine learning approaches have been proposed to learn discriminative patterns directly from neuroimaging data \cite{chandru2025artificial}. However, conventional supervised approaches typically require task-specific training data and can be difficult to develop reliably in small, imbalanced clinical cohorts \cite{adhikari2026dcgan}.

This limitation is especially acute in African clinical contexts, where neuroimaging datasets remain small and unevenly distributed across diagnostic centers \cite{wogu2025fair}. Data are also frequently incomplete, with missing sequences, heterogeneous field strengths, and demographic imbalance across
diagnostic groups \cite{wogu2025labeled}. Under these conditions, training a dedicated deep learning model from scratch for each new clinical task or site is often impractical. Foundation models pretrained on large, predominantly non-African cohorts have been proposed as an alternative: a frozen backbone
paired with a lightweight classifier could, in principle, transfer useful representations to a new population using only a small labeled set \cite{ruffini2026benchmarking}. However, this assumption has rarely been tested outside the well-curated datasets on which models such as BrainIAC \cite{tak2026generalizable} and Neuro-JEPA \cite{huang2026learning} were developed. Whether these learned representations transfer to heterogeneous, resource-constrained clinical data from African populations therefore remains an open question.

In this work, we evaluate whether representations learned by four pretrained brain MRI foundation models transfer to an African clinical cohort under a frozen-backbone, low-data setting. We compare them with a 3D ViT trained from scratch, evaluate performance across T1w, T2w, FLAIR, and multimodal inputs, and examine spatial attribution maps to characterize model behavior beyond aggregate classification metrics. Our findings provide insights into the strengths and limitations of these foundation models, highlighting their applicability and potential challenges in underrepresented clinical populations.

%\section{Related Work}
%\input{2_related}

\section{Method}
\subsection{Dataset and Preprocessing}

We used the Nigerian Brain MRI dataset \cite{wogu2025labeled} which contains a total of 761 MRI sequences collected from 88 participants across three diagnostic centers in Nigeria: 31 participants with dementia ($65.26 \pm 14.11$ years, range: $36$--$86$), 22 participants with Parkinson's disease ($61.14 \pm 14.60$ years, range: $36$--$79$),
and 35 healthy controls ($34.14 \pm 12.75$ years, range: $12$--$75$). Data were acquired on scanners with field strengths ranging from 0.3 T to 1.5 T. Not all participants had all three MRI sequences (T1w, T2w, FLAIR) available. Eleven control participants had one or more missing MRI sequences; consequently, the number of participants varied by sequence: T1w ($n=79$), T2w ($n=80$), T1w+T2w ($n=75$), and FLAIR ($n=80$).

The dataset contains multiple intrasession acquisitions for some subjects. To ensure consistent evaluation, we used BRISQUE (Blind/Referenceless Image Spatial Quality Evaluator) \cite{brisque}, a no-reference image quality assessment metric, to identify and select the acquisition with the lowest BRISQUE score in the evaluation. The dataset was skull-stripped using HD-BET \cite{isensee2019automated}. Additional preprocessing was applied as required by the specific foundation model used for evaluation.

\subsection{Modeling}
We evaluate four pretrained 3D foundation models under a frozen-backbone protocol, alongside ViT3D as a from-scratch baseline. Each model processes the input volume into a sequence of patch tokens that serve as input to the probing head. The pretrained weights are loaded from published checkpoints and remain frozen throughout training. This isolates the contribution of each model's learned representations from the capacity added by the probe.

\noindent{\textbf{NeuroVFM}}\cite{kondepudi2025health} is a ViT-B pretrained on 5.24 million brain CT and MRI volumes using Vol-JEPA, a volumetric JEPA objective. It learns a shared latent representation across CT and MRI through masked volumetric prediction, providing a large-scale multimodal pretraining paradigm for downstream neuroimaging tasks.

\noindent{\textbf{Neuro-JEPA}}\cite{huang2026learning} is a ViT-B pretrained on 1,551,862 multimodal MRI scans (T1, T2, and FLAIR) using a JEPA objective, with mixture-of-experts routing. It produces patch-level representations without a CLS token and uses a sparse MoE architecture to encode multimodal brain MRI.

\noindent{\textbf{BrainIAC}}\cite{tak2026generalizable} is a ViT-B pretrained on 32,015 multiparametric brain MRIs from 16 datasets spanning 10 neurological conditions using SimCLR contrastive learning. It produces 768-dimensional representations from the CLS token for downstream adaptation.

\noindent{\textbf{Primus}}\cite{wald2026primus} is a custom transformer encoder--convolutional decoder architecture pretrained on the OpenMind dataset \cite{openmind} using a Volume Contrastive (VoCo) \cite{voco} learning objective. We use only the pretrained encoder and discard the decoder.

\noindent{\textbf{ViT3D}} serves as the from-scratch baseline and is trained end-to-end rather than frozen.

For single-modality inputs the forward pass is straightforward: the volume goes through the backbone's patch embedding, through the transformer encoder, and outputs a tensor of patch tokens. For multimodal inputs, the first convolutional layer of the patch embedding must accept more than one input channel. The original weights were learned on single-channel data. To extend them to multi-channel inputs, we replicate the weight tensor across the new channels and divide by the channel count. This adaptation is applied for every multimodal configuration before training begins.

\subsection{Linear Probing Protocol}
For each pretrained backbone, we train a linear classifier on the frozen features. The probing head consists of dropout followed by a single linear projection: Dropout($p=0.3$) $\rightarrow$ Linear(hidden\_dim, 3). The hidden dimension depends on the model: 768 for ViT-based backbones (Neuro-JEPA, NeuroVFM, BrainIAC), 864 for Primus, and 384 for ViT3D-B/16. 

All heads use the same training setup. We optimize with AdamW at a learning rate of 0.001 and weight decay of $1 \times 10^{-4}$. The learning rate warms up linearly from 1\% to 100\% over five epochs, then decays following a cosine schedule to 1\% of the initial rate. The loss is cross-entropy with label smoothing of 0.1, which acts as a regularizer against overconfidence. Training runs for a maximum of 50 epochs with a batch size of 4 in a 5-fold cross-validation configuration.

\subsection{Evaluation}
We report five metrics per fold. Accuracy captures overall correctness, while balanced accuracy and the Matthews correlation coefficient (MCC) account for the moderate class imbalance in our cohort. Macro-F1 averages per-class performance without weighting by support, while weighted F1 weights by class size. The macro area under the receiver-operating characteristic curve (AUC-OVR) measures per-class separability independent of any decision threshold, computed in a one-versus-rest fashion. 

Five models are evaluated under identical conditions. Neuro-JEPA \cite{huang2026learning}, NeuroVFM \cite{kondepudi2025health}, BrainIAC \cite{tak2026generalizable}, and Primus \cite{wald2026primus} are used as frozen backbones with a linear probe trained on top, while ViT3D is trained end-to-end from scratch as a within-architecture baseline.

\subsection{Explainability}

We applied gradient-weighted class activation mapping (Grad-CAM) to obtain spatial attribution maps for model predictions. For each subject, gradients of the predicted-class logit with respect to the model's spatial feature representations were used to estimate feature importance, which was projected back to the corresponding patch locations and upsampled to the original voxel resolution. For multimodal inputs, each modality was evaluated independently with the remaining modalities masked to zero, producing modality-specific attribution maps. Subject-level attribution maps were averaged within each class and subsequently min--max normalized.

\section{Experimental Results}
\subsection{Classification performance}
 
Table~\ref{tab:main_results} reports five-fold classification performance across the four modality configurations. ViT3D, trained end-to-end, generally outperforms the frozen foundation-model backbones across the evaluated tasks and metrics. Its strongest overall performance is observed on the T1w+T2w fusion setting, achieving an accuracy of 0.533, balanced accuracy of 0.482, macro-F1 of 0.434, MCC of 0.272, and AUC of 0.672. Neuro-JEPA attains a higher AUC than ViT3D on T1w+T2w (0.676 vs.\ 0.672) and FLAIR (0.684 vs.\ 0.675). On FLAIR, Neuro-JEPA achieves the strongest overall performance among the frozen models, with an accuracy of 0.448, balanced accuracy of 0.409, macro-F1 of 0.324, MCC of 0.128, and AUC of 0.684. The remaining frozen foundation-model backbones generally show weaker classification performance. 

\begin{table*}[h]
\centering
\caption{\textbf{Classification Performance on African Brain MRI.}
Mean over 5 subject-level cross-validation folds. 
\textbf{Bold} = best, \underline{underline} = second best. 
Cell shading indicates relative performance.}
\label{tab:main_results}

% % ============================================================
% % (a) T1w and T2w
% % ============================================================
 \textbf{(a) Single-Modality MRI}

 \vspace{2pt}

 \begin{NiceTabular}{lccccc | ccccc}
 \toprule
 Model
 & \multicolumn{5}{c}{T1w ($n=79$)}
 & \multicolumn{5}{c}{T2w ($n=80$)} \\
 \cmidrule(lr){2-6}\cmidrule(lr){7-11}
 & Acc & BalAcc & F1 & MCC & AUC
 & Acc & BalAcc & F1 & MCC & AUC \\
 \midrule

 BrainIAC\cite{tak2026generalizable}
 & \cellcolor[HTML]{CE7388}0.381
 & \cellcolor[HTML]{CE7388}0.326
 & \cellcolor[HTML]{E58A88}0.207
 & \cellcolor[HTML]{CE7388}-0.022
 & \cellcolor[HTML]{FFF5CC}\underline{0.587}
 & \cellcolor[HTML]{FEE6B9}0.375
 & \cellcolor[HTML]{FEDFB3}0.322
 & \cellcolor[HTML]{F6A895}0.194
 & \cellcolor[HTML]{FFF5CC}-0.025
 & \cellcolor[HTML]{E6F4BC}0.559 \\

 Neuro-JEPA\cite{huang2026learning}
 & \cellcolor[HTML]{CF7488}0.382
 & \cellcolor[HTML]{D77C88}0.331
 & \cellcolor[HTML]{E38788}0.205
 & \cellcolor[HTML]{D27788}-0.017
 & \cellcolor[HTML]{EB928A}0.548
 & \cellcolor[HTML]{F1F9C8}\underline{0.402}
 & \cellcolor[HTML]{F1F9C8}\underline{0.351}
 & \cellcolor[HTML]{FFF8D0}\underline{0.233}
 & \cellcolor[HTML]{E0F2B9}\underline{0.032}
 & \cellcolor[HTML]{FEEBBC}0.515 \\

 NeuroVFM\cite{kondepudi2025health}
 & \cellcolor[HTML]{E58A88}0.392
 & \cellcolor[HTML]{DC8088}0.333
 & \cellcolor[HTML]{CE7388}0.188
 & \cellcolor[HTML]{E48888}\underline{0.000}
 & \cellcolor[HTML]{CE7388}0.535
 & \cellcolor[HTML]{FCC7A3}0.361
 & \cellcolor[HTML]{FFF3C8}0.333
 & \cellcolor[HTML]{DC8088}0.176
 & \cellcolor[HTML]{F9FCD3}0.000
 & \cellcolor[HTML]{DCF0B6}\underline{0.566} \\

 Primus \cite{wald2026primus}
 & \cellcolor[HTML]{E78C88}\underline{0.393}
 & \cellcolor[HTML]{E78C88}\underline{0.339}
 & \cellcolor[HTML]{F7AB96}\underline{0.230}
 & \cellcolor[HTML]{E38788}-0.001
 & \cellcolor[HTML]{FBC09F}0.564
 & \cellcolor[HTML]{CE7388}0.323
 & \cellcolor[HTML]{CE7388}0.278
 & \cellcolor[HTML]{CE7388}0.169
 & \cellcolor[HTML]{CE7388}-0.158
 & \cellcolor[HTML]{CE7388}0.439 \\

 \midrule

 ViT3D
 & \cellcolor[HTML]{73AC91}\textbf{0.507}
 & \cellcolor[HTML]{73AC91}\textbf{0.465}
 & \cellcolor[HTML]{73AC91}\textbf{0.407}
 & \cellcolor[HTML]{73AC91}\textbf{0.241}
 & \cellcolor[HTML]{73AC91}\textbf{0.651}
 & \cellcolor[HTML]{73AC91}\textbf{0.462}
 & \cellcolor[HTML]{73AC91}\textbf{0.407}
 & \cellcolor[HTML]{73AC91}\textbf{0.308}
 & \cellcolor[HTML]{73AC91}\textbf{0.141}
 & \cellcolor[HTML]{73AC91}\textbf{0.634} \\

 \bottomrule
 \end{NiceTabular}

 \vspace{10pt}

  %% ============================================================
 %  (b) Multimodal + FLAIR
  %% ============================================================
 \textbf{(b) Multimodal and FLAIR MRI}

 \vspace{2pt}

 \begin{NiceTabular}{lccccc | ccccc}
 \toprule
 Model
 & \multicolumn{5}{c}{T1w+T2w ($n=75$)}
 & \multicolumn{5}{c}{FLAIR ($n=80$)} \\
 \cmidrule(lr){2-6}\cmidrule(lr){7-11}
 & Acc & BalAcc & F1 & MCC & AUC
 & Acc & BalAcc & F1 & MCC & AUC \\
 \midrule

 BrainIAC\cite{tak2026generalizable}
 & \cellcolor[HTML]{CE7388}0.400
 & \cellcolor[HTML]{CE7388}0.322
 & \cellcolor[HTML]{CE7388}0.190
 & \cellcolor[HTML]{CE7388}-0.043
 & \cellcolor[HTML]{F8AC96}0.543
 & \cellcolor[HTML]{CE7388}0.351
 & \cellcolor[HTML]{CE7388}0.311
 & \cellcolor[HTML]{CE7388}0.173
 & \cellcolor[HTML]{CE7388}-0.090
 & \cellcolor[HTML]{CE7388}0.533 \\

 Neuro-JEPA\cite{huang2026learning}
 & \cellcolor[HTML]{F8AD97}\underline{0.426}
 & \cellcolor[HTML]{F19F90}\underline{0.347}
 & \cellcolor[HTML]{EB928A}\underline{0.218}
 & \cellcolor[HTML]{FCC4A1}\underline{0.039}
 & \cellcolor[HTML]{73AC91}\textbf{0.676}
 & \cellcolor[HTML]{97D1A4}\underline{0.448}
 & \cellcolor[HTML]{AFDDAA}\underline{0.409}
 & \cellcolor[HTML]{E0F2B9}\underline{0.324}
 & \cellcolor[HTML]{9FD5A6}\underline{0.128}
 & \cellcolor[HTML]{73AC91}\textbf{0.684} \\

 NeuroVFM\cite{kondepudi2025health}
 & \cellcolor[HTML]{E88D88}0.413
 & \cellcolor[HTML]{E08488}0.333
 & \cellcolor[HTML]{D37888}0.195
 & \cellcolor[HTML]{EF9A8E}0.000
 & \cellcolor[HTML]{FDC9A4}0.556
 & \cellcolor[HTML]{FAB49A}0.376
 & \cellcolor[HTML]{F5A794}0.333
 & \cellcolor[HTML]{D87D88}0.182
 & \cellcolor[HTML]{FEDEB2}0.000
 & \cellcolor[HTML]{FFF3C8}0.598 \\

 Primus\cite{wald2026primus}
 & \cellcolor[HTML]{E88D88}0.413
 & \cellcolor[HTML]{E08488}0.333
 & \cellcolor[HTML]{D37888}0.195
 & \cellcolor[HTML]{EF9A8E}0.000
 & \cellcolor[HTML]{CE7388}0.510
 & \cellcolor[HTML]{FFF3C8}0.401
 & \cellcolor[HTML]{FEE2B5}0.356
 & \cellcolor[HTML]{FDCFA6}0.242
 & \cellcolor[HTML]{F8FCD2}0.051
 & \cellcolor[HTML]{FED5AA}0.580 \\

 \midrule

 ViT3D
 & \cellcolor[HTML]{73AC91}\textbf{0.533}
 & \cellcolor[HTML]{73AC91}\textbf{0.482}
 & \cellcolor[HTML]{73AC91}\textbf{0.434}
 & \cellcolor[HTML]{73AC91}\textbf{0.272}
 & \cellcolor[HTML]{76B294}\underline{0.672}
 & \cellcolor[HTML]{73AC91}\textbf{0.466}
 & \cellcolor[HTML]{73AC91}\textbf{0.436}
 & \cellcolor[HTML]{73AC91}\textbf{0.411}
 & \cellcolor[HTML]{73AC91}\textbf{0.175}
 & \cellcolor[HTML]{7BBA99}\underline{0.675} \\

 \bottomrule
 \end{NiceTabular}

\end{table*}

\subsection{Per-class performance}
 
Table~\ref{tab:per_class_f1} reports per-class F1 scores on the T1w+T2w configuration. The frozen backbones show a bias toward the Dementia class, with BrainIAC, NeuroVFM, and Primus achieving zero F1 for Parkinson's and very low F1 for Control (0.028, 0.021, and 0.117, respectively). Neuro-JEPA is a partial exception, achieving non-zero F1 for both Control (0.126) and Parkinson's (0.052). In contrast, ViT3D is the only model with a relatively non-trivial F1 across all three classes (Control $0.412$, Dementia $0.629$, Parkinson $0.128$).
 
BrainIAC, NeuroVFM, and Primus predict almost exclusively the Dementia class, while the end-to-end trained ViT3D distributes its predictions across all three diseases, correctly classifying 11/22 Control, 26/31 Dementia, and 8/22 Parkinson's subjects.

% , the only model to achieve non-zero recall on all three classes.

\begin{figure}[t]
\centering
\includegraphics[width=0.95\textwidth]{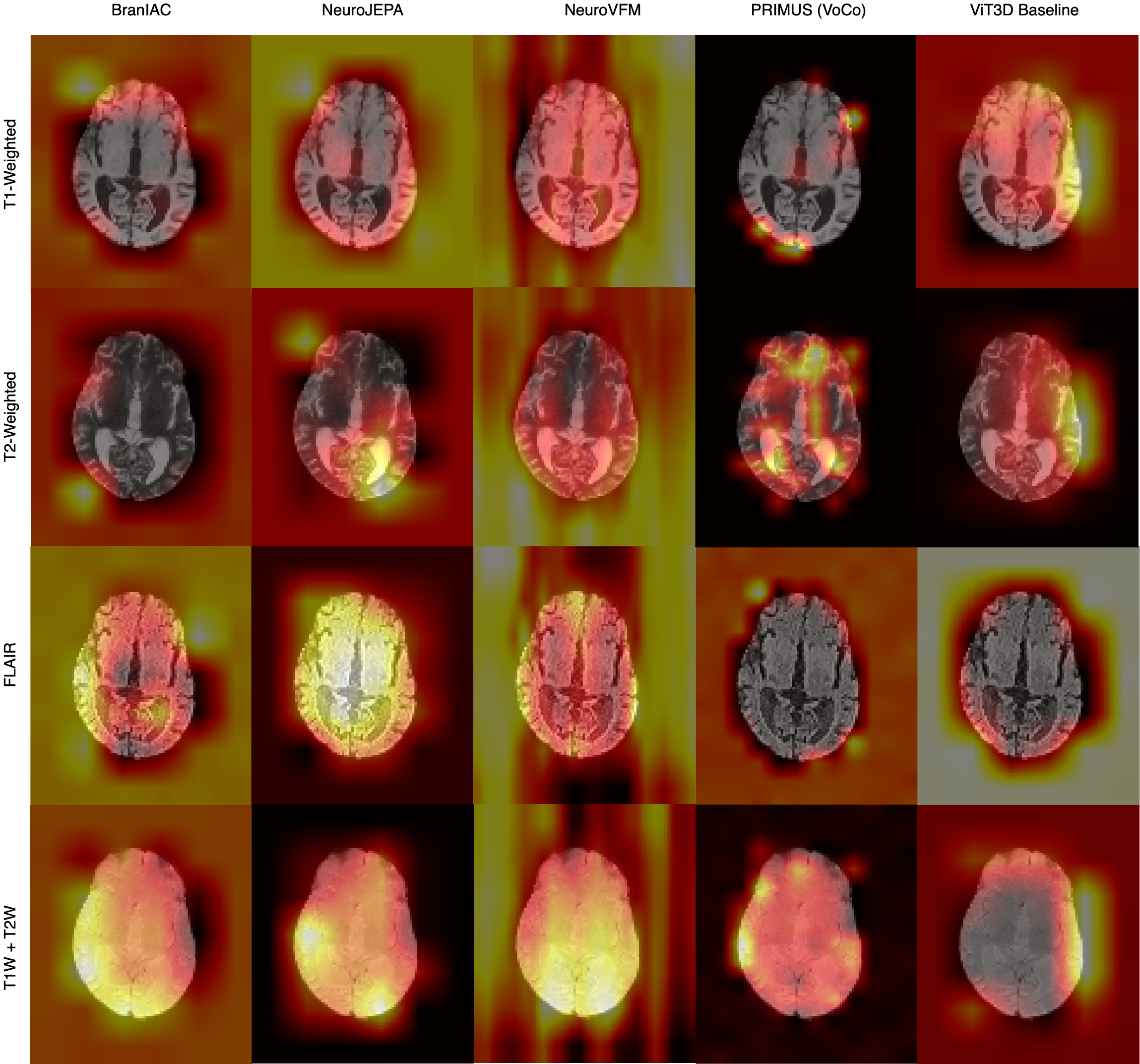}
\caption{Grad-CAM attention maps by modality (rows) and model (columns).}
\label{fig:gradcam}
\end{figure}

\subsection{Explainability Analysis}
To probe what the models are attending to, Figure~\ref{fig:gradcam} shows mean Grad-CAM attention maps per class. The lowest-performing frozen backbones (BrainIAC and NeuroVFM) produce diffuse, largely unstructured activation spanning the entire field of view rather than localizing to the brain. Primus attends to sparse, patchy regions concentrated at the brain periphery rather than deep gray-matter structures. Most notably, ViT3D exhibits a strong, spatially consistent hotspot at one lateral edge of the image across nearly every modality. A shortcut feature may support confident predictions without them being well grounded in pathology, tempering the otherwise favorable results of ViT3D in Table \ref{tab:main_results}. Although the attribution maps suggest possible shortcut learning, they cannot establish causality.

% Table 2: Per-class F1 (macro-averaged over 4 tasks, mean across tasks shown; see note below on aggregation choice)

\begin{table}[t]
\centering
\caption{\textbf{Per-class F1 Score}.}
\label{tab:per_class_f1}
\begin{tabular}{lccc}
\toprule
Model & Control & Dementia & Parkinson \\
\midrule
BrainIAC & 0.028 & 0.545 & 0.000 \\
Neuro-JEPA & 0.126 & 0.557 & 0.052 \\
NeuroVFM & 0.021 & 0.534 & 0.000 \\
Primus & 0.117 & 0.509 & 0.000 \\
ViT3D & \textbf{0.412} & \textbf{0.629} & \textbf{0.128} \\
\bottomrule
\end{tabular}
\end{table}

\subsection{Statistical significance}
 
Paired $t$-tests on 5-fold accuracy (ViT3D vs.\ each frozen backbone) reached significance only on the T1w task (BrainIAC $p=0.044$, Neuro-JEPA $p=0.041$, NeuroVFM $p=0.021$); no other model/task comparison reached $p<0.05$, consistent with the limited statistical power of a 5-fold protocol at this sample size ($n=79$–$80$).

\section{Discussion}
Our findings suggest that reported generalization of brain MRI foundation models does not necessarily extend to small, heterogeneous and clinically distinct cohorts under a frozen-backbone protocol. This is consistent with prior work showing that foundation models do not necessarily yield superior fine-tuned performance compared with standard supervised models in low-data settings, and that direct benchmarking against simpler baselines remains essential \cite{nair2026impact}. In our benchmark, the foundation models did not exhibit uniformly transferable representations across downstream tasks, contrary to the broad transferability often reported in large-scale brain MRI foundation-model studies. 

This pattern likely reflects a combined effect of both the model and the data. On the model side, using purely frozen features with a simple classification head may not allow sufficient adaptation to local disease phenotypes, scanner variability, or domain shift in this cohort. On the data side, the dataset is small, asymmetric, and non-representative, which increases fold-to-fold variability and weakens the reliability of fixed representations. These observations align with the broader literature, which shows that neuroimaging foundation models often benefit from large-scale training and carefully matched downstream adaptation. For example, BrainIAC was introduced as a self-supervised foundation model for brain MRI designed to support broad downstream use, but its success still depends on how well the downstream setting matches pretraining conditions \cite{tak2026generalizable}. Our results reinforce the view that frozen foundation models are not automatically sufficient in low-resource settings unless domain adaptation and dataset-specific considerations are taken into account.

Neuro-JEPA was the most interesting special case in our benchmark. It performed relatively well on FLAIR and showed improved behavior when T1 and T2 were combined, suggesting that the learned representation may make more effective use of complementary sequence information in this cohort. This is especially plausible for FLAIR, which often highlights pathology-related differences more clearly than conventional T1 or T2 imaging. However, these gains were not uniform, indicating that even strong multimodal objectives do not fully overcome scanner shift, cohort heterogeneity, and limited sample size.

\section{Conclusion}
In this study, we evaluated four pretrained brain MRI foundation models under a frozen-backbone protocol on an African clinical cohort comprising dementia, Parkinson's disease, and control participants. Across T1w, T2w, FLAIR, and T1w+T2w inputs, the frozen foundation models generally failed to outperform a 3D ViT trained from scratch, with several models exhibiting uneven class-wise performance. Neuro-JEPA was a notable exception, achieving competitive AUC on multimodal and FLAIR inputs.

These findings challenge the prevailing assumptions that frozen foundation model backbones can universally adapt to low-data environments. The poor transferability observed is likely driven by severe domain shift: models pretrained predominantly on data from cohorts outside the target population struggle to interpret the heterogeneous acquisition protocols, lower-field scanner noise (0.3 T to 1.5 T), and distinct disease etiology profiles characteristic of African clinical settings. These findings underscore the need for neuroimaging foundation models to incorporate geographically and clinically diverse pretraining data.

\section{Limitations}
Our study has important limitations. The dataset is small and heterogeneous. The models were evaluated as frozen feature extractors with a simple classification head, which may underuse the pretrained representations. We also did not evaluate lightweight adaptation strategies such as LoRA fine-tuning on the target population, directly assess potential shortcut features contributing to confident predictions, or quantify the data and computational costs of calibrating these models for underrepresented populations. Future work should investigate parameter-efficient fine-tuning, shortcut-feature robustness, population-specific calibration, and richer multimodal fusion, particularly with T1+T2+FLAIR, to determine whether the observed performance limitations arise primarily from the data, the frozen setup, or both.

\begin{credits}
% \subsubsection{\ackname} None.

\subsubsection{\discintname}

The authors have no competing interests to declare that are relevant to the content of this article.

\end{credits}
% \newpage

\bibliographystyle{splncs04}
\bibliography{biblio}





% LNCS Homepage, \url{http://www.springer.com/lncs}, last accessed 2023/10/25
% \end{thebibliography}

\end{document}